# Automatic Detection of Reuses and Citations in Literary Texts

J.-G. Ganascia[1,3], P. Glaudes[2,3] and A. Del Lungo[4]

[1] ACASA team, LIP6 - University Pierre and Marie Curie, Paris, France

[2] Littérature française, XIX$^e$-XXI$^e$ siècles, Université Paris-Sorbonne, Paris, France

[3] Labex OBVIL, PRES Sorbonne University, Paris, France

[4] ALITHILA, Université Charles de Gaulle, Lille 3, Lille, France

**Contact author**: Jean-Gabriel Ganascia, ACASA team,

LIP6 – University Pierre et Marie Curie, B.C. 169, 4, place Jussieu, 75005, Paris, France

email: Jean-Gabriel.Ganascia@lip6.fr

# Automatic Detection of Reuses and Citations in Literary Texts

J.-G. Ganascia[1,3], P. Glaudes[2,3] and A. Del Lungo[4]

[1] ACASA team, LIP6 - University Pierre and Marie Curie, Paris, France

[2] Littérature française, XIX[e]-XXI[e] siècles, Université Paris-Sorbonne, Paris, France

[3] Labex OBVIL, PRES Sorbonne University, Paris, France

[4] ALITHILA, Université Charles de Gaulle, Lille 3, Lille, France

## 1.     Introduction

The question of influence in arts in general and in literature in particular is certainly one of the most difficult and challenging problems that humanists have to deal with. As Harold Bloom expresses in his classical essay "The Anxiety of Influence" (Bloom, 1997), whose title is so evocative, there is a tension between the artist's aspiration to originality and his/her references to the tradition within which his/her work is rooted. Like a teenager, the artist is both continuing and disrupting a tradition. Very often, this double tendency is revealed by more or less explicit references to the works of the recognized forerunners that may be deferent, ironic or even irreverent, as in parodies. For more than forty years now, modern theories of literature (Compagnon, 1979) insist on the role of paraphrases, rewritings, citations, reciprocal borrowings and mutual contributions of any kinds. The notions of *intertextuality* (Kristeva, 1969; 1974), *transtextuality*, *hypertextuality/hypotextuality* (Genette, 1982), were introduced in the seventies and eighties to approach these phenomena. The careful analysis of these references is of particular interest in evaluating the distance that the creator voluntarily introduces with his/her masters. This not only helps us to understand the surrounding context of the authors and the factors that motivated them to create, but also to comprehend their goals and strategies. Obviously, it's not a question of plagiarism, which is the fraudulent and masked use of existing material, but, on the contrary, a process of differentiation and of intentional distinction that the detection of references helps to explicit.

With the digitization of contents, it now becomes possible to automatically detect some of these different phenomena (Coffee & al., 2012). However, references to others' writings are often so altered that it's difficult to recognize their original source. So, we have to identify inexact rewritings, which is far more difficult than simple duplications, because the number of possible alterations is huge and their form surprising. There are many works attempting to identify these approximate textual homologies. Some of them are based on the distribution of words or on the density of n-grams (e.g. (Allen & al., 2010), (Horton & al., 2010), (Büchler & al., 2010)). Others are based on plagiarism detection techniques, for instance on the use of 'fingerprinting'.

However, the importance is in the goal of the detection. Some researchers look for statistical characterizations of authors' lexicon allowing to track influences. For instance, where two fragments of text share more frequently words or series of words than is usual, this could be seen as an indicant of influence from one to the other. Our goal here, with the PHŒBUS project, is somehow different: it is to detect the reuses, citations, borrowings etc. through huge quantities of texts, not with the aim of statistically quantifying the influences, but to enlighten particular cases of borrowings. For instance, we would like to detect cases where one author (e.g. Balzac) reuses another one's text (e.g. Gautier), to cast light on the recycling of scientific writings, to look at the origin of the shared citations (e.g. citation of Shakespeare in French literature) and the way they are introduced, to enumerate the uses of proverbs in novels, etc. For us, the question is not to find a number that quantifies the degree of borrowing, but to perform their qualitative analysis and the subsequent interpretation. Ultimately, our aim would be to use this work as a basis for a hypertext that could associate to a literary work the intellectual landscape around it, with annotated links that would explain the nature and the rationale of the association. This would contribute to making understood the "pretext" of the text, the word "pretext" being both taken with the etymological meaning of what precedes a text, and with the current acceptation that corresponds to what motivates it.

Through the PHŒBUS project, computer scientists from the Computer Science Laboratory of the University Pierre and Marie Curie (LIP6-UPMC) are collaborating with the literary teams of Paris-

Sorbonne University and Lille 3 University in order to develop efficient tools for literary studies; the latter take advantage of modern computer science techniques to detect borrowings on huge masses of texts and to help to put them in context. Our ultimate goal is to detect approximate homologies between French literature of the 19$^{th}$ century and the newspapers of the same epoch. This will soon become possible with the *Europeana Newspaper* project (cf. http://www.europeana-newspapers.eu/) to digitize the press of second half 19$^{th}$ century; the Europeana Newspaper project attempts to digitize 18 millions of journal pages, among which 10 millions will be available in full text. By restricting the corpus to a single language, for instance to French or to German, this corresponds to 1 or 2 millions of pages, i.e. roughly few tens of Gigabytes. We are far below from what is currently called "big data" by companies like Google or Facebook, which face terabytes, or indeed petabytes, but this quantity of data is not negligible and requires taking great care of the algorithmic complexity of the procedures.

In this context, we have developed a piece of software that automatically detects and explores networks of textual homologies in classical literature. Written in PROLOG this program has been extensively tested on many texts, for instance on Isidore Ducasse texts (Lautréamont, 2009) that are known to contain many reuses and on "La comédie humaine" (Balzac, 1976-1981) from Honoré de Balzac, which, according to (Duclos, 2013), reuses some texts of his friend Théophile Gautier (Gautier, 2002). We consider our approach more appropriate for our purpose, which is to detect significant textual homologies and then to facilitate their interpretation, than other similar approaches, e.g. (Roe, 2012) (Büchler, Crane, Mueller, Burns, & Heyer, 2011) (Coffee & al., 2012), because it allows the splicing of the recurring n-grams and the filtering of their combinations, according to syntactic or semantic criteria.

This paper describes the principles on which our program is based, the results that have already been obtained and the prospective for the near future. The remaining is divided into four parts. The first recalls the distinction between various types of borrowings such as plagiarism, pastiches or citations. The second enumerates the criteria that are retained to characterize textual homologies and citations on which we are focusing here. A third part describes the implementation of our program and shows its efficiency by comparison with manual detection. Finally, we show some of the results that have already been obtained with the PHŒBUS program.

## 2.     Distinctions

Before going into the details of the description of the implemented techniques, let us distinguish the different figures of *literary transformations*, including *textual reuses* and *citations*, which we aim to automatically detect, from two similar notions, the *plagiarism* and the *imitation*. According to Gérard Genette (1979), the purpose of *transtextuality* or of *textual transcendence* is to establish links between texts and consequently to study the above-mentioned phenomena among which he differentiates *imitations* from *transformations*, and then, among transformations, *parodies*, *travesties* and *transpositions*, while among imitations, he discerns *pastiches*, *charges* and *forgery* (Genette, 1982). The propose of our research is to identify different types of textual transformation by comparing huge quantities of texts, while imitation and plagiarism are out of the scope of the present study.

To be more specific, let us recall that plagiarism consists in stealing the work of another, i.e. in fraudulently appropriating his/her texts, without mentioning explicitly their origin. As such, the plagiarism is considered as an unethical practice that has to be tracked down and prosecuted. Many techniques have been developed to detect plagiarism, considered as a plague, because intellectual work is stolen (Potthast & al., 2010; 2011). By contrast, the pastiche and other sorts of imitation, such as the charge or the forgery, are artistic practices that replicate an artist, a style or a period. There appears to be nothing wrong with this, except that charges or pastiches may mock or criticize well-known authors. Many well-known writers, for instance Marcel Proust, began by pastiches both for fun and to improve their style. The forgery pretends to be the continuation of an author's work after his/her death, as, for instance, the following of Homer's Iliad by more recent authors. This could only be condemned when the signature is erroneous, i.e. when by contrast with plagiarism, its author doesn't mention himself, but the authorities whom he intends to imitate. The detection of imitation is close to the identification of literary style (Dinu, Niculae, & Sulea, 2012), which requires capturing the essence of an artist's style or of a period.

Halfway from detection of plagiarisms and identification of imitation, the recognition of transformations like textual reuses and citations helps to track the literary influences, and allows penetrating the spirit of the epoch. Some of the textual reuses and citations are conscious, other not. They may correspond to explicit – or implicit – and more or less distorted quotations. Usually, textual reuses and other type of transformations proceed by altering a piece of text, while citations are verbatim, but this is not always the case. Whichever phenomenon it is, when a sufficient part of the original text is kept, its fragments can be recognized. This is exactly what we attempt to do automatically here. Reuses and approximate citations are far more difficult to detect than plagiarism, because the original fragments of text may be distorted, but far less than with pastiches and other figures of imitation. Their detection is of great interest for researchers concerned with *intertextuality* and *transtextuality*. They are useful in literary criticism, because the way the original text is transformed greatly informs about the author's mood and intentions.

## 3. Criteria

As previously said, text reuse and citation discovery is inspired from plagiarism detection, but it has to take into account all the alterations that may have transformed the initial text. To specify the type of distortions that affect a text, we started from a hand made study realized by Tania Duclos. She shows in (Duclos, 2013) [cf. *table 1*] how some parts of the *Human Comedy* (Balzac, 1976-1981) reuse fragments of text from Théophile Gautier. As we shall see these manually annotated fragments are very useful because they render possible an evaluation of the efficiency of our algorithm and an optimal adjustment of the parameters.

| *Béatrix* – Honoré de Balzac | *Jenny Colon (Portraits)* – Théophile Gautier |
|---|---|
| « Si elle pouvait par un artifice quelconque porter le costume dues anci temps où les femmes avaient des corsets pointus à échelles de rubans s'élançant minces et frêles de l'ampleur étoffée des jupes en brocatelle à plis soutenus et puissants, s'entouraient de fraises godronnées cachaient leurs bras dans des manches à crevés et à sabots de dentelles d'où la main sortait comme une fleur de sa capsule, et qui rejetaient leurs les mille boucles de leur chevelure sur leurs épaules au delà d'un chignon ficelé de pierreries, elle lutterait avec avantage cont avec les beautés les plus célèbres que vous voyez vêtues ainsi dit-elle en montrant un tableau à Calyste, ### debout, devant un tenant une m un papier et chantant avec un seigneur brabançon, pendant qu'un nègre verse dans un verre à patte du vieux vin d'Espagne et qu'une la vieille femme de charge arrange des biscuits. » | « Les costumes romanesques de Piquillo conviennent beaucoup au type de beauté de Mlle Colon; les grandes robes de lampas ou de brocatelle aux plis soutenus et puissants, les hautes fraises godronnées et frappées à l'emporte-pièce, comme on en voit dans les dessins de Romain de Hooge; les manches à crevés et à sabots de dentelles, dont la main sort comme le pistil du calice d'une fleur, les feutres à ganse de perles, à plumes crespelées, les chaînes et les rivières de diamants écaillant d'étincelles papillotantes la blancheur mate de la poitrine, les corsets pointus à échelles de rubans s'élançant minces et frêles de l'ampleur étoffée des jupes: - toute la toilette abondante et fantasque du seizième siècle s'adapte merveilleusement à la physionomie de Mlle Colon, que l'on prendrait, dans un de ses costumes capricieux, pour un de ces belles dames des gravures d'Abraham Bosse, qui marchent gravement une tulipe à la main, suivies du petit page nègre qui porte leur queue, leur chien et leur manchon, dans les allées bordées de buis d'un parterre du temps de Louis XIII. » |

*Table 1: example of hand coded comparison* (Duclos, 2013) *between a fragment of Béatrix* (Balzac, 1976-1981) *on the left and a fragment of Théophile Gautier* (Gautier, Portraits contemporains, 1874) *on the right.*

While looking in detail at table 1, it appears that some of the passages highlighted are identical, while others are somehow different. For instance *"en brocatelle à plis soutenus et puissants, s'entouraient de fraises godronnées"* becomes *"de brocatelle aux plis soutenus et puissants, les hautes fraises godronnées"* and *"des manches à crevés et à sabots de dentelles d'où la main sortait comme une fleur de sa capsule"* becomes *"les manches à crevés et à sabots de dentelles, dont la main sort comme le pistil du calice d'une fleur"*. Some fragments look far more difficult to identify, because they are composed of isolated words or even different words (e.g. *"diamants"* and *"pierreries"* or *"tableau"* and *"gravures"*) of similar meanings. Here, we try to detect string homologies where some words may be missing, especially stop words, i.e. articles, pronouns or prepositions. It may also happen that the number and the gender of nouns or adjectives and the persons or the tenses of verbs have changed as when *"d'une main mignonne frappée de fossettes"* is transformed in *"des mains mignonnes frappées de fossettes"* or when *"la main sortait comme une fleur de sa capsule"* becomes *"la main sort comme le*

*pistil du calice d'une fleur"*. Finally, as this example shows, some words may disappear or, contrarily, appear during the transformation.

## 4. Detection of Fragments with Holes

*Algorithm*

Among efficient existing plagiarism detection techniques, many are based on fingerprints built with the hash coding of character strings (Potthast, Eiselt, Barron-Cedeño, Stein, & Rosso, 2011), (Potthast, Stein, Barron-Cedeño, & Rosso, 2010), (Burrows, Tahaghoghi, & Zobel, 2006). Other techniques evaluate the statistical distribution of vocabulary with a vector space model of the texts and a cosine similarity measure that quantifies their closeness. However these don't seem to be appropriate to our purpose, because they require significant chunks of texts to operate meaningfully, while we attempt to detect small recurrent pieces (e.g. of four or five words) that correspond to proverbs or expressions.

We have implemented and adapted the fingerprinting method and we have evaluated it on the reuses isolated by Tania Duclos (Duclos, 2013). It helped us to optimize the values of the different parameters. The process was divided into four steps:

1. Preparation of the text using natural language processing techniques
2. Extraction of elementary recurring sequences of words
3. Splicing the elementary recurring sequences into taller recurring sequences
4. Filtering the resulting sequences

The first step was divided into two sub-steps: the first eliminates the "stop words", i.e. articles, prepositions, pronouns, auxiliary verbs, etc.; the second makes use of the Snowball (Porter, 2001; Tomlinson, 2004) stemmer so as to reduce the words to their root, which allows being independent from the inflected forms used in the text. For instance, the words *"fishing", "fished", "fish", "fishes"* and *"fisher"* are reduced to the same root word *"fish"*.

The second step consists in extracting elementary recurring sequences of words characterized by their minimal size, i.e. by the minimal number of consecutive non-"stop words" they contain, which we call the window size — noted $n_w$ —. In addition, in order to allow missing words, we introduce possible holes. This means that a window of size 4 does not necessarily correspond to 4 consecutive words. The maximum number of possible holes — noted $n_h$ — constitutes a parameter for our algorithm. It means that window size $n_w$ of 3 and a number of holes $n_h$ of 2 covers the following sequences: $M_1 M_2 M_3 - M_1 M_2 M_4 - M_1 M_3 M_4 - M_1 M_3 M_5 - M_1 M_4 M_5 - M_1 M_2 M_5$ where $M_1, M_2, M_3, M_4$ and $M_5$ are five consecutive non-"stop words".

Once the similar fragments are discovered, they are adjoined end to end, i.e. spliced, which build taller blocs of recurring sequences. This third step allows the detection of big recurring pieces of text.

Lastly, we contrast what we call "weak words", which are neither very significant nor very uncommon, to "strong words" that are less common, and more informative than weak ones. These words may be given manually or extracted automatically using a probability distribution of words. The fourth step of our algorithm filters the blocs of similar words of which the number of "strong words" is bigger than a minimal threshold, for instance 4. This allows eliminating noise, without loosing information, because only segments with significant words are retained. This filtering also ensures that the recurrences are longer than a minimal limit; for instance, if the threshold is equal to 4 with $n_w = 3$ and $n_h = 2$, it means that the pattern is necessarily composed of at least two consecutive elementary fragments.

*Evaluation of the algorithm*

We have tested our algorithm by comparing the obtained results with the similarities noted by Tania Duclos in her thesis (Duclos, 2013), except the questionable ones like those that are highlighted in yellow in *table 1*. As usual in information retrieval, we have computed the classical evaluation criteria that are the *precision*, or the accuracy of the answers, the *recall*, which corresponds to the percentage of retrieved annotated examples, and the *F-score accuracy*. More precisely, we have reckoned

$$F_\beta = (1-\beta^2) \cdot \frac{precision \cdot recall}{\beta^2 \cdot precision + recall}$$

with $\beta$ = 0.5. This calculation allows getting the optimal values of the parameters (cf. *table 2*).

Relying on the results presented in *table 2*, we have chosen the following parameters: $n_w$ = *3* and $n_h$ = 2, which appears to be an optimal tradeoff between recall and precision. However, depending on the problem, the parameters may vary.

The program has been implemented in SWI-Prolog (cf. (SWI-Prolog's home)) using an external table to store hash-coded texts. This program is quite efficient: for instance it takes less than 10 minutes to index all the Balzac's "Human Comedy" (Balzac, 1976-1981), which contains more than 25 millions of characters, on a 2GHz MacPro. It then takes a couple of minutes to discover text reuses on entire novels.

### *Interface*

As previously mentioned, our goal is to develop a tool that helps scholars to interpret textual reuses and borrowings, which requires putting them in place by inserting them in their environment. We are not only interested in the transformed fragments of the original text, but also in the context of the original and altered pieces of text, because it's only when seen in their particular setting that transformations make sense.

| Recall | $N_w=1$ | $N_w=2$ | $N_w=3$ | $N_w=4$ | $N_w=5$ | $N_w=6$ |
|---|---|---|---|---|---|---|
| $N_t=0$ | 0.91 | 0.9 | 0.89 | 0.56 | 0.25 | 0.13 |
| $N_t=1$ | 0.91 | 1 | 0.89 | 0.67 | 0.5 | 0.25 |
| $N_t=2$ | 0.91 | 0.9 | 0.89 | 0.78 | 0.5 | 0.25 |
| $N_t=3$ | 0.91 | 0.9 | 1 | 0.89 | 0.5 | 0.38 |
| $N_t=4$ | 0.91 | 0.9 | 0.89 | 0.67 | 0.38 | 0.38 |
| $N_t=5$ | 0.91 | 0.8 | 0.78 | 0.56 | 0.63 | 0.63 |

| Precis. | $N_w=1$ | $N_w=2$ | $N_w=3$ | $N_w=4$ | $N_w=5$ | $N_w=6$ |
|---|---|---|---|---|---|---|
| $N_t=0$ | 0.53 | 1 | 1 | 1 | nd | nd |
| $N_t=1$ | 0.51 | 0.78 | 1 | 1 | 1 | 1 |
| $N_t=2$ | 0.51 | 0.57 | 1 | 1 | 1 | 1 |
| $N_t=3$ | 0.63 | 0.45 | 0.69 | 1 | 1 | 1 |
| $N_t=4$ | 0.63 | 0.43 | 0.65 | 0.88 | 1 | 1 |
| $N_t=5$ | 0.64 | 0.37 | 0.5 | 0.5 | 0.53 | 1 |

| F-score | $N_w=1$ | $N_w=2$ | $N_w=3$ | $N_w=4$ | $N_w=5$ | $N_w=6$ |
|---|---|---|---|---|---|---|
| $N_t=0$ | 0.57 | 0.98 | 0.98 | 0.86 | nd | nd |
| $N_t=1$ | 0.56 | 0.81 | 0.98 | 0.91 | 0.83 | 0.63 |
| $N_t=2$ | 0.56 | 0.62 | 0.98 | 0.95 | 0.83 | 0.63 |
| $N_t=3$ | 0.67 | 0.5 | 0.73 | 0.98 | 0.83 | 0.75 |
| $N_t=4$ | 0.67 | 0.48 | 0.68 | 0.82 | 0.75 | 0.75 |
| $N_t=5$ | 0.68 | 0.41 | 0.54 | 0.51 | 0.55 | 0.89 |

*Table 2: comparison of the results obtained for the different $n_w$ and $n_h$*

To understand at a glance the results obtained by our algorithm, we need to visualize them and restore them in their surroundings. That is why we have designed a Graphical User Interface (GUI) that shows the textual transformations at two levels of granularity. On the first level, the zones of similarities, i.e. the detected reuses, are plotted on a square diagram in which each axis corresponds to one text (cf. figure 1). By clicking on each red zone, corresponding to a detection textual reuse, we get access to the second level which presents the environment of the detected textual reuse that is highlighted in color (cf. figure 2).

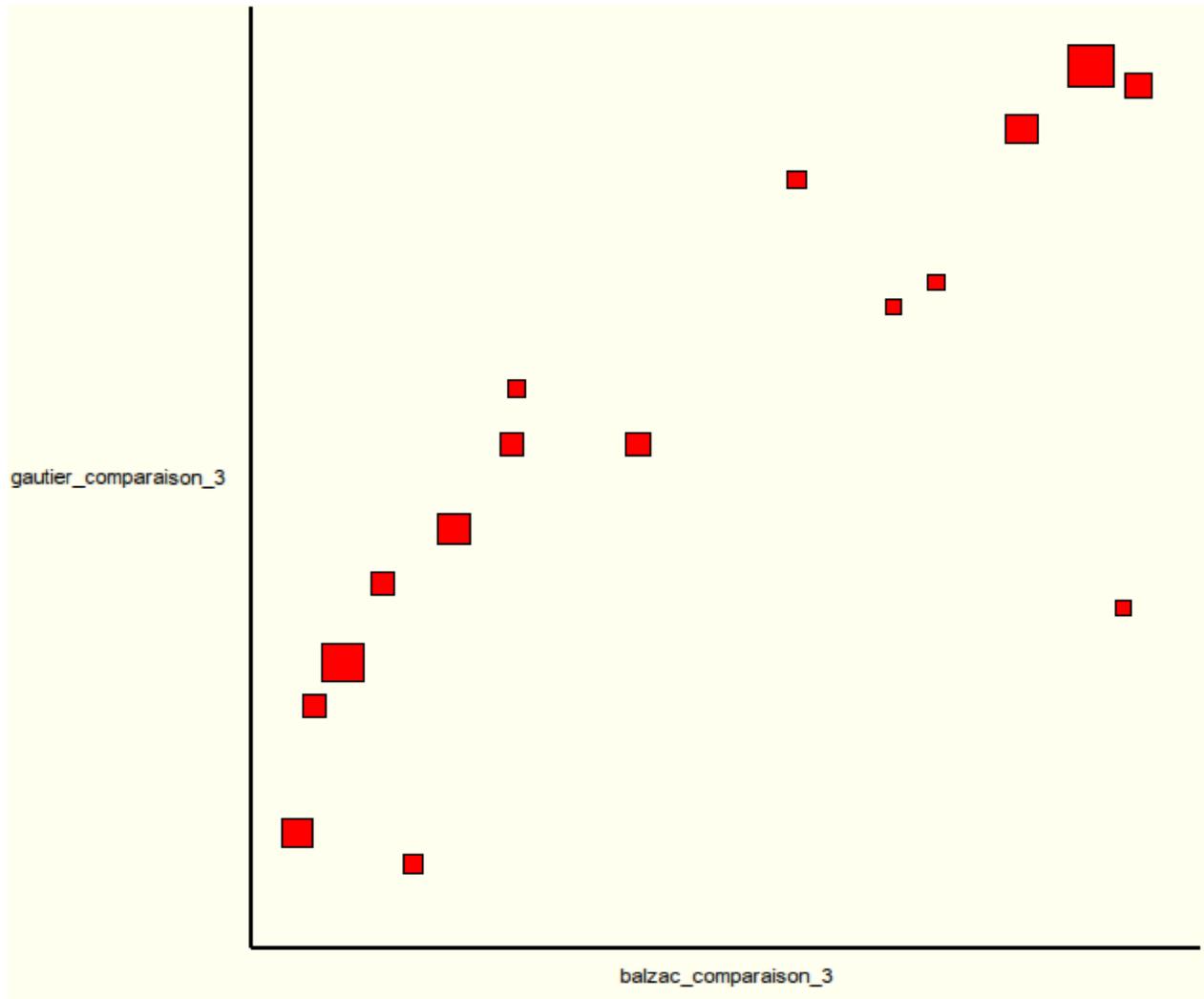

*Figure 1: zones of similarities between fragments of Gautier (y-axe) and fragments of Balzac (x-axe)*

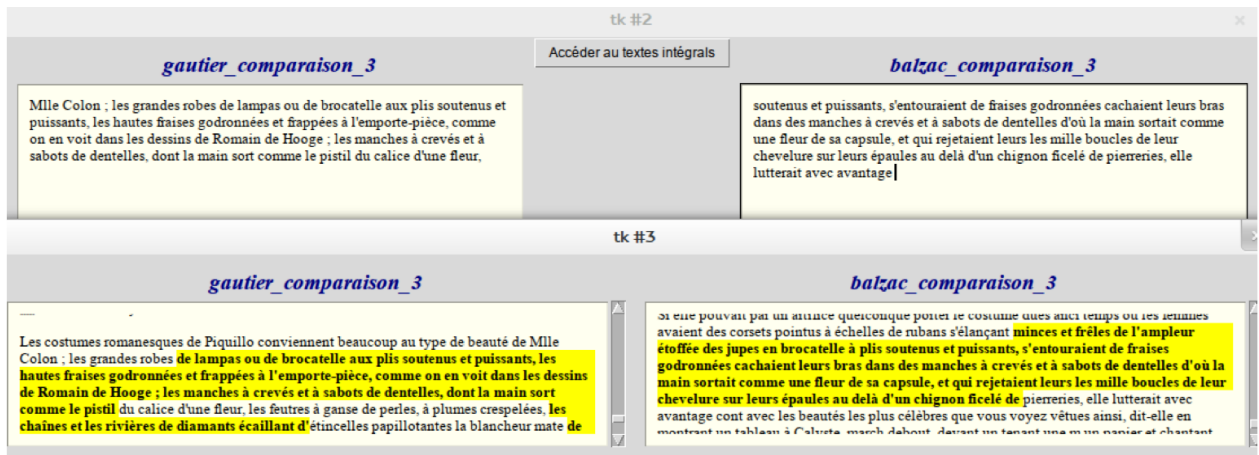

*Figure 2: a detected textual reuse with its environment*

## 5. Results

Using this program, we were able to retrieve all the hand-coded reuses of (Duclos, 2013), except the "yellow" one (see *table 1*). We have also detected many other interesting citations and reuses, for instance a reuse of the Gautier's Novel entitled "Mademoiselle de Maupin" (Gautier, 2002) in the Balzac's Novel "Modeste Mignon" (Balzac, 1976-1981), which had not been mentioned before, or a citation of George Lyttleton both in "Delphine" (de Staël, 1869) and in "Ursule Mirouët" (Balzac, 1976-1981).

We have also tested the system comparing Lautréamont's work (Lautréamont, 2009) and Buffon, which gave surprising results. For instance we have detected the following reuse:

*"du bec supérieur s'élève une caroncule charnue, de forme conique et sillonnée par des rides transversales assez profondes." (Buffon)*

*"ou encore, comme la caroncule charnue, de forme conique, sillonnée par des rides transversales assez profondes, qui s'élève sur la base du bec supérieur du dindon" (Lautréamont)*

We then compared Lautréamont to the French moralists, e.g. Pascal, La Rochefoucauld or La Bruyère. We have retrieved many textual reuses, among which were some interesting distortions, for instance Pascal's aphorism *"Nous naissons injustes; car chacun tend à soi: cela est contre tout ordre."* that has been rewritten in *"Nous naissons justes. Chacun tend à soi. C'est envers l'ordre"*. Or, La

Rochefoucauld's maxim LXXVIII "*L'amour de la justice n'est, en la plupart des hommes, que la crainte de souffrir l'injustice.*" that has been altered by Lautréamont in "*L'amour de la justice n'est, en la plupart des hommes, que le courage de souffrir l'injustice.*"

We have picked up many of the textual transformations mentioned by Gérard Genette in (Genette, 1982) which were automatically retrieved. For instance, confronting *Les plaideurs* of Jean Racine to *Le Cid* of Corneille, we have retrieved that the following verse of Pierre Corneille (Le Cid): *"Ses rides sur son front ont gravé ses exploits"* has been transformed in *"Ses rides sur son front gravaient tous ses exploits."* (Les plaideurs, Racine).

The software may also detect parodies as shown in the comparison of "Le chapelain décoiffé" from Nicolas Boileau with "Le cid" written by Pierre Corneille (cf. table 3).

| *Le Cid* - Pierre Corneille | *Le chapelain décoiffé* – Nicolas Boileau |
|---|---|
| Ô rage ! ô désespoir ! ô vieillesse ennemie !<br>N'ai-je donc tant vécu que pour cette infamie ?<br>Et ne suis-je blanchi dans les travaux guerriers<br>Que pour voir en un jour flétrir tant de lauriers ?<br>À des partis plus hauts ce beau fils doit prétendre ;<br>Et le nouvel éclat de votre dignité<br>Lui doit enfler le cœur d'une autre vanité. | Ô rage ! ô désespoir: ô perruque ma mie !<br>N'as-tu donc tant vécu que pour cette infamie?<br>N'as-tu trompé l'espoir de tant de perruquiers,<br>Que pour voir en un jour flétrir tant de lauriers?<br>À de plus hauts partis, Phlipote peut prétendre !<br>Et le nouvel éclat de cette pension<br>Lui doit bien mettre au cœur une autre ambition ! |

*Table 3: comparison of "Le Cid" (Pierre Corneille) with "Le chapelain décoiffé" (Nicolas Boileau)*

Since many people mention the influence of the phrenology and the physiognomy on Balzac's work, we have tried to detect similarities between Balzac's novels and scientific books. We have obtained many interesting results among which one is presented in table 4.

| *La Vieille Fille,* « Pléiade », vol. IV, p. 857 –Balzac | *La Physiognomonie et la Phrénologie*, Chap. XV (« *Des signes tirés de l'examen du nez* »), pp. 224-225. - Bourdon |
|---|---|
| Son nez aquilin contrastait avec la petitesse de son front, car il est rare que cette forme de nez n'implique pas un beau front. Malgré de grosses lèvres rouges l'indice d'une grande bonté, ce front annonçait trop peu d'idées pour que le cœur fût dirigé par l'intelligence : elle devait être bienfaisante sans grâce. | Le nez et le front sont presque toujours dans un accord parfait; ce que l'un d'eux annonce l'autre le confirme unanimes sont leurs décisions. Il est rare qu'un nez ignoble soit uni à un beau front intellectuel. Tel nez, tel front, tel esprit cette règle a peu d'exceptions. |

*Table 4: comparison of Balzac novels with scientific texts of physiognomy*

Lastly, our software has revealed some very interesting and substantial textual reuses that had not been mentioned before by the literary scholars who participated to the previous publications of Balzac's works in (Balzac, 1976-1981). One of them has been obtained by the confrontation of the different novels of the Human Comedy against themselves. In *Madame Firminani,* published in 1832, it shows the recycled description of a person drawn from *La pathologie de la vie sociale,* printed earlier, in 1830 (cf. table 5). From this point of view, this constitutes a true discovery.

| *La pathologie de la vie sociale* - Balzac | *Madame Firminani* – Balzac |
|---|---|
| Mais il est une personne dont la voix harmonieuse imprime au discours un charme également répandu dans ses manières. Elle sait et parler et se taire; s'occupe de vous avec délicatesse; ne manie que des sujets de conversation convenables, ses mots sont heureusement choisis; son langage est pur, sa raillerie caresse et sa critique ne blesse pas. Loin de contredire avec l'ignorante assurance d'un sot, elle semble chercher, en votre compagnie, le bon sens ou la vérité. Elle ne disserte pas plus qu'elle ne dispute, elle se plaît à conduire une discussion, qu'elle arrête à propos. D'humeur égale, son air est affable et riant, sa politesse n'a rien de forcé, son empressement n'est pas servile; elle réduit le respect à n'être plus qu'une ombre douce; elle ne vous fatigue jamais et vous laisse satisfait d'elle et de vous. Entraîné dans sa sphère par une puissance inexplicable, vous retrouvez son esprit de bonne grâce empreint sur les choses dont elle s'environne : tout y flatte la vue, et vous y respirez comme l'air d'une patrie. Dans l'intimité, cette personne vous séduit par un ton naïf. Elle est naturelle. Jamais d'effort, de luxe, d'affiche. Ses sentiments sont simplement rendus parce qu'ils sont vrais. Elle est franche sans offenser aucun amour-propre. Elle accepte les hommes comme Dieu les a faits, pardonnant aux défauts et aux ridicules; concevant tous les âges et ne s'irritant de rien, parce qu'elle a le tact de tout prévoir. Elle oblige avant de consoler; elle est tendre et gaie, aussi l'aimerez-vous irrésistiblement. Vous la prenez pour type et lui vouez un culte. | Avez-vous, pour votre bonheur, rencontré quelque personne dont la voix harmonieuse imprime à la parole un charme également répandu dans ses manières, qui sait et parler et se taire, qui s'occupe de vous avec délicatesse, dont les mots sont heureusement choisis, ou dont le langage est pur ? Sa raillerie caresse et sa critique ne blesse point; elle ne disserte pas plus qu'elle ne dispute, mais elle se plaît à conduire une discussion, et l'arrête à propos. Son air est affable et riant, sa politesse n'a rien de forcé, son empressement n'est pas servile; elle réduit le respect à n'être plus qu'une ombre douce; elle ne vous fatigue jamais, et vous laisse satisfait d'elle et de vous. Sa bonne grâce, vous la retrouvez empreinte dans les choses desquelles elle s'environne. Chez elle, tout flatte la vue, et vous y respirez comme l'air d'une patrie. Cette femme est naturelle. En elle, jamais d'effort, elle n'affiche rien, ses sentiments sont simplement rendus, parce qu'ils sont vrais. Franche, elle sait n'offenser aucun amour-propre; elle accepte les hommes comme Dieu les a faits, plaignant les gens vicieux, pardonnant aux défauts et aux ridicules, concevant tous les âges, et ne s'irritant de rien, parce qu'elle a le tact de tout prévoir. À la fois tendre et gaie, elle oblige avant de consoler. Vous l'aimez tant, que si cet ange fait une faute, vous vous sentez prêt à la justifier. Vous connaissez alors madame Firmiani. |

*Table 5: comparison of "La pathologie de la vie sociale" (Honoré de Balzac) with "Madame Firminani" (Honoré de Balzac)*

## 6. Perspectives

In the near future, we shall extensively use our system in many fields of literature, especially on 19[th] century French literature, with Balzac's work. This was the original aim of the PHŒBUS project funded by the CNRS and the OBVIL laboratory. More precisely, PHŒBUS is intended to investigate the textual reuses in different Balzac's works, especially, but not only, the novels of the "Human Comedy" and between Balzac's works and his contemporaries' works like Théophile Gautier, Benjamin Constant, George Sand etc. We also plan to digitalize the journals where many authors have published articles either under their own names or anonymously and to compare them with the "Human Comedy". A further application will be to extensively compare the work of Balzac with the scientists of his time, especially with physicians, phrenologists and physiognomists. Furthermore, we will conduct a thorough comparison with similar approaches, in particular with the algorithms developed in the PhiloLogic project (Allen & al., 2010), in (Büchler & al., 2011) and in the Tesserae project (Coffee & al., 2012). Lastly, we will attempt to test it on big quantities of texts, for instance, on the $19^{th}$ century press. The first evaluation of the current results of the PHŒBUS program with Balzac's work shows that it is possible to extend it to one or two higher orders of magnitude, which corresponds to our evaluation of the size of the $19^{th}$ century press.

We are also developing the interface to make it publicly available to scholars as free software, available through the web.

## Funding

This work has been conducted within the LABEX OBVIL project, and received financial state aid managed by the Agence Nationale de la Recherche, as part of the programme "Investissements d'avenir" under the reference ANR-11-IDEX-0004-02. It was also supported by the CNRS through the project Phoebus (PEPS)

## 7. References

**Allen T., Cooney C., Douard S., Horton R., Morrissey R., Olsen M., Roe G., Voyer R.** (2010), "Plundering Philosophers: Identifying Sources of the Encyclopédie", Journal of the Association for History and Computing, vol. 13, no. 1, Spring 2010, URL http://hdl.handle.net/2027/spo.3310410.0013.107 (last access 7 November 2013)

**Bloom, H.** (1997). *The Anxiety of Influence: A Theory of Poetry*, Oxford University Press.

**Büchler, M., Crane, G., Mueller, M., Burns, P., & Heyer, G.** (2011). One Step Closer To Paraphrase Detection On Historical Texts: About The Quality of Text Re-use Techniques and the Ability to Learn Paradigmatic Relations. (G. K. Thiruvathukal, & S. E. Jones, Éds.) *Journal of the Chicago Colloquium on Digital Humanities and Computer Science* .

**Balzac, H.** (1976-1981). *La comédie humaine* (Vol. I-XII). (C. L. Pléiade, Éd.) Paris: Gallimard.

**Büchler, M., Geßner, A., Heyer, G., Eckart, T.** (2010): Detection of Citations and Text Reuse on Ancient Greek Texts and its Applications in the Classical Studies: eAQUA Project. Digital Humanities 2010 Conference, London, URL http://dh2010.cch.kcl.ac.uk/academic-programme/abstracts/papers/pdf/ab-637.pdf, last access 11 November 2013.

**Büchler, M., G. Crane, M. Mueller, P. Burns, and G. Heyer (2011).** One Step Closer To Paraphrase Detection On Historical Texts: About The Quality of Text Re-use Techniques and the Ability to Learn Paradigmatic Relations. in Thiruvathukal, G. K. & Jones, S. E., (Éds). *Journal of the Chicago Colloquium on Digital Humanities and Computer Science*.

**Burrows, S., Tahaghoghi, S., & Zobel, J.** (2006). Efficient Plagiarism Detection for Large Code Repositories. *Software - Practice and Experience , 37*, 151-175.

**Coffee, N., Koenig, J.-P., Poornima, S., Forstall, C., Ossewaarde, R., & Jacobson, S.** (2012). Intertextuality in the Digital Age. *Transactions of the American Philological Association , 142* (2).

**Coffee, N., Koenig, J.-P., Poornima, S., Forstall, C., Ossewaarde, R., & Jacobson, S.** (2012, July). The Tesserae Project: Intertextual Analysis of Latin Poetry. *Litterary and Linguistic Computing* , 1-8.

**Compagnon, A.** (1979), *La Seconde main ou le travail de la citation*, Seuil, Paris.

**Dinu, L. P., Niculae, V., & Sulea, O.-M.** (2012). Pastiche detection based on stopword rankings. Exposing impersonators of a Romanian writer. *EACL 2012 - Workshop on Computational Approaches to Deception Detection* (pp. 72-77). Avignon: Association for Computational Linguistics.

**Duclos, T.** (2013). L'intertextualité dans une Fille d'Eve et Béatrix d'Honoré de Balzac. Paris-Sorbonne University.

**Forstall, C., Jacobson, S. and Scheirer, W.**, (2011), "Evidence of Intertextuality: Investigating Paul the Deacon's *Angustae Vitae*," Literary and Linguistic Computing, September 2011, Vol. 26, No. 3, September 2011.

**Gautier, T.** (1874). *Portraits contemporains.* Paris, France: Charpentier et Cie.

**Gautier, T.** (2002). *Romans, contes et nouvelles* (Vol. I-II). (C. d. Pléiade, Éd.) Paris: Gallimard.

**Genette, G.** (1979), *Introduction à l'architexte*, Seuil, coll. « Poétique », Paris.

**Genette, G.** (1982), *Palimpsestes : La Littérature au second degré*, Seuil, coll. « Essais », Paris.

**Horton, R., Olsen, M., and Roe, G.** (2010) "Something Borrowed: Sequence Alignment and the Identification of Similar Passages in Large Text Collections." *Digital Studies/ Le champ numérique* 2.1. Available at: http://www.digitalstudies.org/ojs/index.php/ digital_studies/article/view/190/235. (last access 7 November 2013).

**Kristeva, J.** (1969). *Sémiotikè, Recherches pour une sémanalyse,* col. "Tel Quel", Editions du Seuil.

**Kristeva, J.** (1974). *La Révolution du langage poétique,* col. "Tel Quel", Editions du Seuil.

**Lautréamont.** (2009). *Œuvres complètes.* (C. d. Pléiade, Éd.) Paris: Gallimard.

**Porter, M.** (2001, October). *Snowball: A language for stemming algorithms*. Consulté le October 22, 2012, sur Snowball: http://snowball.tartarus.org/texts/introduction.html

**Potthast, M., Eiselt, A., Barron-Cedeño, A., Stein, B., & Rosso, P.** (2011). Overview of the 3rd International Competition on Plagiarism Detection. Dans V. Petras, P. Forner, & P. D. Clough (Éd.), *Notebook Papers of CLEF 11 Labs and Workshops*.

**Potthast, M., Stein, B., Barron-Cedeño, A., & Rosso, P.** (2010). An Evaluation Framework for Plagiarism Detection. Dans C.-R. Huang, & D. Jurafsky (Éd.), *23rd International Conference on Computational Linguistics (COLING 10)*, (pp. 997-1005). Stroudsburg, Pennsylvania.

**Roe, G. R.** (2012). Intertextuality and Influence in the Age of Enlightenment: Sequence Alignment Applications for Humanities Research . *Digital Humanities.* Hamburg.

*SWI-Prolog's home*. (s.d.). Consulté le Octobre 22, 2012, sur SWI-Prolog: http://www.swi-prolog.org/

**Staël (de), G.** (1869). *Delphine.* Paris: Garnier frères.

**Tomlinson, S.** (2004). Lexical and Algorithmic Stemming Compared for 9 European Languages with Hummingbird SearchServer TM at CLEF 2003. Dans C. Peters (Éd.), *Working Notes for the CLEF 2003 Workshop.* Springer